\pdfoutput=1

\documentclass[11pt]{article}

\usepackage[]{EACL2023}

\usepackage{times}
\usepackage{latexsym}

\usepackage[T1]{fontenc}

\usepackage[utf8]{inputenc}

\usepackage{microtype}

\usepackage{inconsolata}
\usepackage{amsmath}
\usepackage{cleveref}
\crefformat{section}{\S#2#1#3} 

\usepackage{graphicx}
\usepackage{booktabs}
\usepackage{multirow}
\usepackage{algorithm}
\usepackage{algorithmic}

\usepackage{ascii}

\usepackage{newfloat}
\usepackage{listings}
\usepackage{xcolor}
\usepackage{arydshln}
\usepackage{colortbl}

\definecolor{jred}{RGB}{196, 38, 11}
\definecolor{jblue}{RGB}{41, 52, 190}
\definecolor{jgreen}{RGB}{18, 141, 21}

\definecolor{wxjiao}{RGB}{18, 21, 141}

\definecolor{alizarin}{rgb}{0.82, 0.1, 0.26}

\definecolor{xing}{rgb}{1.0, 0.03, 0.0}

\definecolor{jinhui}{rgb}{1,0.6,0}

\title{
Scaling Back-Translation with Domain Text Generation \\for Sign Language Gloss Translation}

\author{
Jinhui Ye$^1$\thanks{\ \ Jinhui Ye and Wenxiang Jiao contributed equally to this work. Work was mainly done when Jinhui Ye was interning at Tencent AI Lab.}  \quad Wenxiang Jiao$^{2}$$^*$ \quad Xing Wang$^{2}$\thanks{\ \ Xing Wang is the corresponding author.}  \quad Zhaopeng Tu$^2$\\
$^1$The Hong Kong University of Science and Technology (Guangzhou) \\ {\asciifamily \normalsize\tt jye624@connect.hkust-gz.edu.cn} \\
$^2$Tencent AI Lab \\  {\asciifamily \normalsize\tt \{joelwxjiao,brightxwang,zptu\}@tencent.com} \\
}

\begin{document}
\maketitle
\begin{abstract}

Sign language gloss translation aims to translate the sign glosses into spoken language texts, which is challenging due to the scarcity of labeled gloss-text parallel data. Back-translation (BT), which generates pseudo parallel data by translating in-domain spoken language texts into sign glosses, has been applied to alleviate the data scarcity problem. However, the lack of large-scale high-quality in-domain spoken language text data limits the effect of BT. In this paper, to overcome the limitation, we propose a \textbf{P}rompt based domain text \textbf{Gen}eration  (\textsc{PGen}) approach to produce the large-scale in-domain spoken language text data. Specifically, \textsc{PGen} randomly concatenates sentences from the original in-domain spoken language text data as prompts to induce a pre-trained language model (i.e., GPT-2) to generate spoken language texts in similar style. Experimental results on three benchmarks of sign language gloss translation in varied languages demonstrate that BT with spoken language texts generated by \textsc{PGen} significantly outperforms the compared methods. In addition, as the scale of spoken language texts generated by \textsc{PGen} increases, the BT technique can achieve further improvements, demonstrating the effectiveness of our approach. We release the code and data for facilitating future research in this field\footnote{Code and data are available at \url{https://github.com/Atrewin/PGen}.}.

\end{abstract}

\section{Introduction}
Sign language is the dominant form of communication for the deaf and hearing impaired community. Sign language processing has received substantial attention in the last few years and achieved significant progress~\cite{bragg2019sign,yin2021including,mtsummit-2021-international-2,de2022machine}. 
Among them, sign language translation (SLT) aims to transform continuous sign language videos into natural spoken language texts~\cite{bungeroth2004statistical,camgoz2018neural}. SLT consists of two sub-tasks: (1) sign language understanding task that recognizes the continued videos to the sign glosses; and (2) spoken language gloss translation~(SLGT) task that generates the spoken language text of the given sign glosses. 
In this work we focus on the second sub-task, i.e., SLGT.

\begin{figure}[t!]
    \centering
	\begin{minipage}{0.80\linewidth}
		\vspace{3pt}
		\centerline{\includegraphics[width=\textwidth]{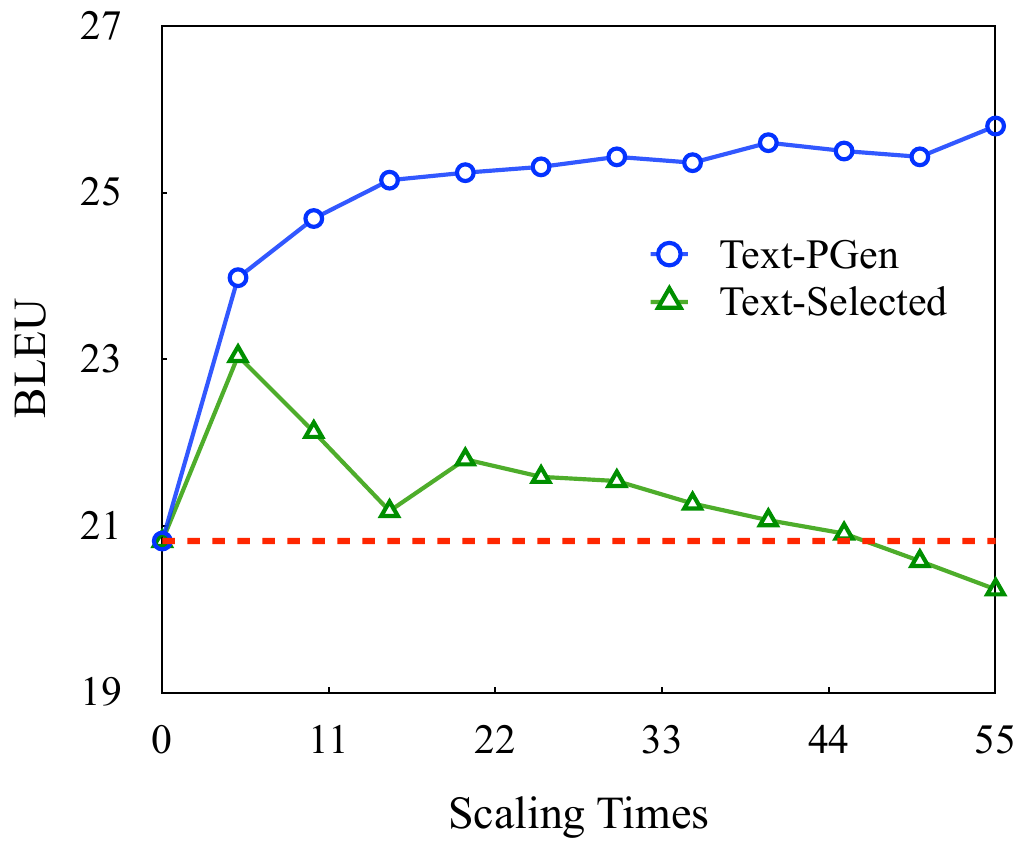}}
	\end{minipage}
	\caption{The translation performance of back-translation when scaling the generated spoken language data from 1) our \textsc{PGen} which uses the large pretrained language model,  and 2) selected in-domain data from large-scale spoken language corpus. The {\color{red} red dashed line} denotes the baseline model without back-translation. Best viewed in color. }
	\label{fig:valid-bleu-vs-scale}
\end{figure}

Data scarcity has been considered the major limitation of sign language gloss translation~\cite{moryossef2021data,zhang2021approaching}.  To alleviate the data scarcity problem, back-translation~\cite{sennrich2016improving}, which translates in-domain spoken language texts into sign glosses
to construct synthetic parallel data, has been adopted and achieved certain success in SLGT.
However, the lack of large-scale high-quality in-domain spoken language text data limits the capability of back-translation for the SLGT task~\cite{zhang2021approaching}. 
The common practice is adopting  data selection~\cite{axelrod2011domain} or  data mining~\cite{jiang2009mining} approaches to obtain in-domain data.  But theses approaches assume the availability of enough in-domain data or expert knowledge (e.g., language background to mine the in-domain data in a specific language), which prevents them from applying to gloss translation task that has broader scenarios (e.g., more languages and domains).

In this work, we propose a \textbf{P}rompt-based domain text \textbf{Gen}eration (\textsc{PGen}) to generate large-scale high-quality in-domain spoken language text data, motivated by the advances in data augmentation with pretrained language models (PLMs).
The main idea is to induce the large PLMs to mimic the style of original spoken language texts with prompt-based learning techniques~\cite{radford2019language,liu2021pre}. Our \textsc{PGen} approach is able to generate large-scale in-domain spoken language texts based on the small-scale original monolingual texts and maintains diversity (\cref{sec:approach-PGen}).
Besides, our approach can be performed without requiring large-scale in-domain data or expert knowledge of the sign language domain and maintain the high quality of generated in-domain spoken language texts. 
Finally, we employ a sequence-sequence pretrained model (e.g, mT5) to translate spoken language texts generated by \textsc{PGen} into sign glosses and synthesize gloss-text pseudo-parallel data  (\cref{sec:approach-BT}).

We conduct extensive analyses of the spoken language texts generated by \textsc{PGen}. We find that the generated and original spoken language texts share a similar word distribution (\cref{sec:domain_text_generation}). To further verify the effectiveness of \textsc{PGen}, we also conduct back-translation experiments for the SLGT
task with large-scale in-domain spoken language  texts generated by \textsc{PGen}  (\cref{sec:sign_lagnauge_glsss_translation}). Experimental results on three widely used benchmark datasets across languages and domains show that back-translation with spoken language texts generated by \textsc{PGen} significantly outperforms the compared methods. 
Most importantly, as shown in Figure~\ref{fig:valid-bleu-vs-scale}, when scaling the spoken language texts generated by \textsc{PGen} in BT approach, the performance of the gloss-to-text translation task can achieve constant improvement while conventional data selection  approach failed.  The contributions of our work are summarized as follows:
\begin{itemize}
\item We propose a novel text generation approach, i.e., \textsc{PGen}, to produce large-scale in-domain spoken language texts which share similar linguistic properties as the original spoken language texts.

\item We scale back-translation with the proposed \textsc{PGen} approach and achieve significant and consistent improvements on three benchmark SLGT datasets.

\item We release the code and the large-scale synthetic gloss-text datasets produced by the proposed approach to promote the research in sign language gloss translation field.
\end{itemize}

\section{Methodology}

\begin{figure*}[t!]
    \centering
    \includegraphics[height=0.4\textwidth]{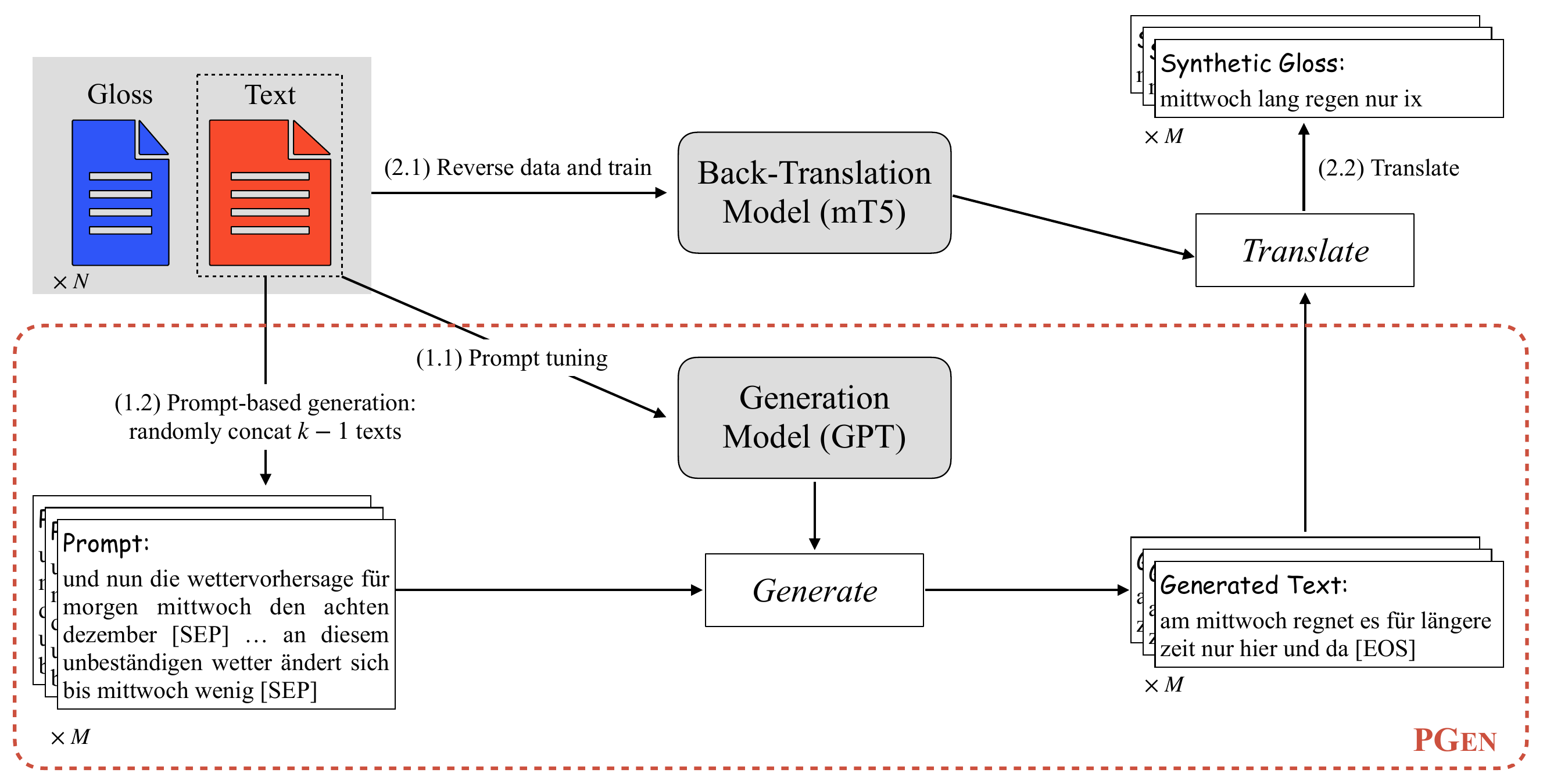}
    \caption{The overall framework of back-translation for gloss-to-text translation in this work. Procedures framed in the {\color{jred!100}  red dashed box} corresponds to our prompt-based domain text generation~(\textsc{PGen}) approach. }
    \label{fig:framework}
\end{figure*}

The whole framework of this work includes two components, i.e., the \textsc{PGen} method for in-domain text generation and the BT model (i.e., text-to-gloss translation) for constructing synthetic data. 
We will introduce more details for these two components in this section. For clarity, we provide the definition of SLGT (i.e., gloss-to-text translation task) and notations used throughout the paper as below. 

\paragraph{Task Definition.}
Let $X$ and $Y$ denote the gloss annotations and spoken language texts, and $\mathcal{X}$ and $\mathcal{Y}$ represent the sentence sets of corresponding languages.
The dataset of gloss-text pairs can be expressed as $\mathcal{D}_{\rm g2t} = \{({\bf x}^i, {\bf y}^i)\}_{i=1}^{N}$, where ${\bf x}^i\in \mathcal{X}$ is the annotations, ${\bf y}^i\in \mathcal{Y}$ is the spoken language sentence and $N$ is the number of pairs. 
Given a sequence of gloss annotations, the task is to output the corresponding fluent and semantically equivalent sentence.

\subsection{Prompt-Based Domain Text Generation}
\label{sec:approach-PGen}

We exploit the large PLMs for domain text generation.
Large PLMs have been successfully applied for text generation in NLP, such as text classification~\cite{kumar2020data} and medical dialogue summarization~\cite{chintagunta2021medically}, for two advantages: (1) PLMs are demonstrated to memorize the knowledge of their training data~\cite{carlini2020extracting}, which usually covers different domains. With proper guidance (e.g., prompts), we can export the memorized sentences that belong to the same domain as the sign language text data. 
(2) Large PLMs are also  able to generate abundant new sentences rather than only the memorized sentence in training data~\cite{qiu2020pre}. 
Therefore, we propose the \textbf{P}rompt-based domain text \textbf{Gen}eration~(\textsc{PGen}) approach to produce large-scale in-domain spoken language text corpora based on the text part of the original small-scale dataset  $\mathcal{D}_{\rm g2t}$.

For the original text $\mathcal{D}_{\rm t} = \{ {\bf y}^i\}_{i=1}^{N}$ from $\mathcal{D}_{\rm g2t}$, and a PLM $\mathcal{M}_{G}$, we attempt to generate an in-domain spoken language text dataset $\hat{\mathcal{D}}_{\rm t} = \{ {\bf \hat{y}}^i\}_{i=1}^{M}$ with a much larger data size than $\mathcal{D}_{\rm t}$ (i.e., $M \gg N$).
As shown in Figure~\ref{fig:framework}, our \textsc{PGen} approach includes two phases:
\begin{itemize}
    \item \textbf{Prompt Tuning:} Following~\cite{kumar2020data}, we finetune the PLM on the original small-scale spoken text dataset $\mathcal{D}_{\rm t}$ with artificial prompts. Specifically, we randomly concatenate $k$ sentences from  $\mathcal{D}_{\rm t}$ as a training sample, i.e., $[{\bf y}^{i_{1}};{\tt [SEP]};{\bf y}^{i_{2}};{\tt [SEP]};\dots;{\bf y}^{i_{k}};{\tt [EOS]}]$, where ${\tt [SEP]}$ and ${\tt [EOS]}$ represent the delimiter and the end-of-sentence tokens, respectively. We denote the finetuned PLM as $\mathcal{M}_{G_{\rm FT}}$. 
    
    \item \textbf{Prompt-Based Generation:} In the generation phase, we randomly select $k-1$ sentences from $\mathcal{D}_{\rm t}$ to form a prompt, i.e., $\tt prompt =  [{\bf y}^{j_{1}};{\tt [SEP]};{\bf y}^{j_{2}};{\tt [SEP]};\dots;{\bf y}^{j_{k-1}};{\tt [SEP]}]$. Then, we input the prompt into $\mathcal{M}_{G_{\rm FT}}$ to generate the $k$-th sentence ${\bf \hat{y}}^{j_k} = \mathcal{M}_{G_{\rm FT}}({\tt prompt})$. We complete the text generation process when the model produces an ${\tt [EOS]}$ token. 
\end{itemize}
According to the design of prompts, the number of permutations for any $k-1$ sentences from the full set with $N$ sentences is $A_N^{k-1} = \frac{N!}{(N-k-1)!} \gg N$, which allows us to generate a large number of in-domain sentences and maintain the  diversity of $\hat{\mathcal{D}}_{\rm t}$.

\subsection{Back-Translation}
\label{sec:approach-BT}

Generally, the BT model is trained on the same dataset for the gloss-to-text task but in the opposite direction, i.e., $\mathcal{D}_{\rm t2g} = \{({\bf y}^i, {\bf x}^i)\}_{i=1}^{N}$.
However, the data scale of $\mathcal{D}_{\rm t2g}$ is too small to develop a well-performing text-to-gloss translation model.
Previous study~\cite{hoang2018iterative} on machine translation also suggests that the quality of BT models heavily affects the performance of the final models. Therefore, we take advantage of pretrained sequence-to-sequence models by finetuning them on $\mathcal{D}_{\rm t2g}$ to improve the performance of the BT model.
Specifically, we utilize a multilingual pretrained model, i.e., mT5~\cite{xue2020mt5}, to support the different languages (i.e., German, Chinese and English) of the SLT benchmarks.

\subsection{Overall Framework}

The workflow of our approach is illustrated in Figure~\ref{fig:framework}, which is divided into four steps: (1.1) finetune the pretrained GPT-2 model on the original small-scale spoken language text dataset; 
(1.2) apply the finetuned GPT-2 model for in-domain spoken language texts generation with artificial prompts; 
(2.1) finetune the pretrained sequence-to-sequence model (i.e., mT5) on the original text-gloss dataset $\mathcal{D}_{\rm t2g}$ to obtain the BT model; (2.2) adopt the BT model to translate the generated in-domain spoken language texts into glosses to synthesize a large-scale pseudo-parallel data, which are combined with the original small-scale dataset $\mathcal{D}_{\rm g2t}$ to train the final gloss-to-text translation model.

\section{Experiments}

In this section, we conduct both intrinsic and extrinsic evaluations~\cite{kumar2020data} for the proposed \textsc{PGen} approach.
For intrinsic evaluation~(\cref{sec:domain_text_generation}), we perform analyses on the domain properties of spoken language texts generated by \textsc{PGen}. As for extrinsic evaluation~(\cref{sec:sign_lagnauge_glsss_translation}), we conduct sign gloss translation experiments using back-translation approach with the generated texts.
The performance of the downstream task can indirectly reflect the effectiveness of \textsc{PGen}.

\subsection{Experimental Setup}

\paragraph{Dataset.}
We employ three widely used benchmark datasets for sign language translation, namely, Phoenix2014T~\cite{camgoz2018neural}, CSL-Daily\footnote{\url{http://home.ustc.edu.cn/~zhouh156/dataset/csl-daily/}}~\cite{zhou2021improving}, and ASLG-PC12\footnote{\url{https://github.com/kayoyin/transformer-slt}}~\cite{othman2012english}, which are in German, Chinese and English, respectively. Statistics of the datasets are presented in Table~\ref{tab:data}.

\begin{table}[t!]
    \centering
    \resizebox{\columnwidth}{!}{
    \begin{tabular}{l c r}
        \toprule
        \bf Dataset & \bf Language Pair &  \bf Gloss-Text Pairs \\
         \midrule
         Phoenix2014T   & DSL-German & 7,086 / 519 / 642 \\
         CSL-Daily  & CSL-Chinese & 18,401 / 1,077 / 1,176 \\
         ASLG-PC12  & ASL-English  & 82,709 / 4,000 / 1,000\\
        \bottomrule
    \end{tabular}
    }
    \caption{Statistics of the three benchmark datasets for gloss-to-text translation used in this work. The third column presents the number of gloss-text pairs in the training, validation and test sets, respectively.}
    \label{tab:data}
\end{table}

\paragraph{Model.}
As shown in Figure~\ref{fig:framework}, there are three kinds of models involved in this work for 1) in-domain text generation, 2) gloss-to-text translation and 3) back-translation, respectively. Details of the model and training settings can be found in~\cref{sec:domain_text_generation} , ~\cref{sec:sign_lagnauge_glsss_translation} and Appendix~\ref{sec:hyper}.

\subsection{Domain Text Generation}
\label{sec:domain_text_generation}

We perform intrinsic evaluation by analyzing the domain properties of spoken language texts generated by \textsc{PGen}. 
Unless otherwise stated, we primarily conduct the analyses on the German Phoenix2014T dataset.
More results on ASLG-PC12 and CSL-Daily datasets can be found in Appendix~\ref{sec:ex-intrinsic}. 

We first adopt the pre-trained GPT-2 model (e.g., German GPT-2\footnote{\url{https://huggingface.co/dbmdz/german-gpt2}} for Phoenix2014T)
and finetune the model on the artificial prompts created from the text part of SLT dataset. Specifically, the PLM is finetuned to predict the next token in the exact way that GPT-2 was pretrained, with the same training procedure and hyper-parameters. Then, we use the finetuned PLM to generate in-domain spoken language texts for each SLT task~(\cref{sec:approach-PGen}). By default, we set the hyper-parameter $k$ to 20. 

For comparison, we consider four types of spoken language texts:
\begin{itemize}
    \item \textsc{Text-Authentic}: The text side of the SLT dataset.
    \item \textsc{Text-PGen}: The spoken language texts generated by our \textsc{PGen} approach with $k=20$. 
    \item \textsc{Text-Selected}: We use the cross-entropy difference selection method~\cite{moore2010intelligent} to collect in-domain texts data from IWSLT17 Multilingual Task\footnote{\url{https://sites.google.com/site/iwsltevaluation2017/TED-tasks}} based on the \textsc{Authentic} texts.
    \item \textsc{Text-General}: We randomly sample sentences from the target text part of IWSLT17 Multilingual Task. \textsc{Text-General} can be considered as the general domain spoken language texts.
\end{itemize}

We measure the similarity of the other three spoken language texts to the \textsc{Text-Authentic} at both word level and sentence level.

\begin{figure}[t!]
    \centering
	\begin{minipage}{0.80\linewidth}
		\centerline{\includegraphics[width=\textwidth]{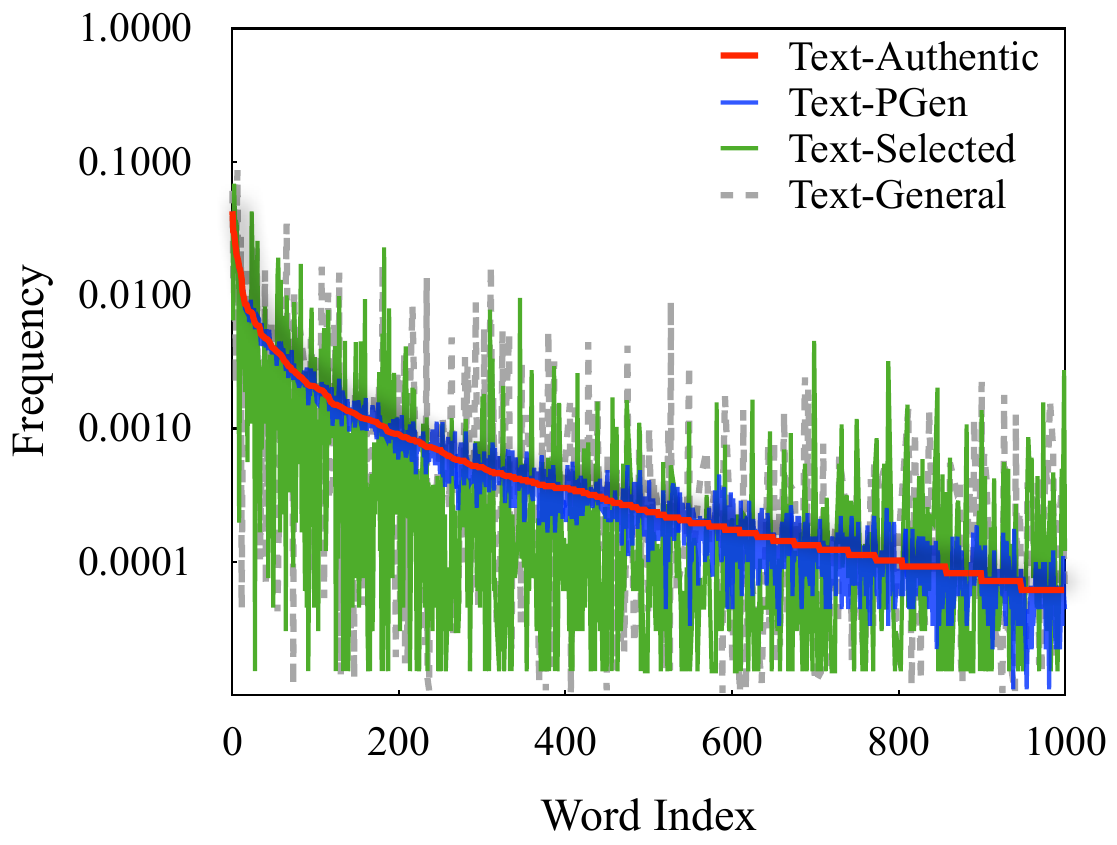}}
	\end{minipage}
	
	\caption{The word frequency distribution on different types of spoken language texts. The X-axis represents different words, while the Y-axis represents the normalized word frequency. Best viewed in color. Analyses is conducted on the German Phoenix2014T}
	\label{fig:WFD}
\end{figure}

\paragraph{Word Distribution.}
\label{Word_Distribution}
Recent study by \citet{wang2022understanding} suggests that the word frequency distributions can reflect the domain difference of datasets. Thus, we count the word frequencies for the above four spoken language texts and present the results by both visualization and the Jensen-Shannon (JS) divergence~\cite{lin1991divergence}. For a fair comparison, we sample the same number of texts from the other three spoken language texts as \textsc{Text-Authentic}~(i.e., 7086 for Phoenix2014T).

Figure~\ref{fig:WFD} visualizes the word frequency distributions of top 10000, in which the words are ranked according to their frequencies in the \textsc{Text-Authentic} corpus. We can observe that \textsc{Text-PGen} ({\color{blue}blue line}) shows a similar distribution as \textsc{Text-Authentic} ({\color{red}red line}) while \textsc{Text-Selected} and \textsc{Text-General} differ from \textsc{Text-Authentic} significantly. This result qualitatively shows that the spoken language texts generated by our \textsc{PGen} approach are more close to the domain of \textsc{Text-Authentic} (e.g. the domain of sign language).

To quantitatively measure the distance between these distributions, we compute the JS divergence expressed as:
\begin{equation}
\small
\begin{split}
 \mathrm{JS}\left (P || Q \right ) = \frac{1}{2} \left (
\mathrm{KL} (P || \frac{P+Q}{2} ) + \mathrm{KL} (Q || \frac{P+Q}{2})
\right ), \nonumber
\end{split}
\end{equation}
where $\mathrm{KL}(\cdot||\cdot)$ denotes the Kullback–Leibler divergence~\cite{kullback1951information} of two distributions (i.e., $P$ and $Q$).
Table~\ref{tab:divergence} lists the JS divergence from the other three corpora to \textsc{Text-Authentic}. We find that the JS divergence from \textsc{Text-PGen} to \textsc{Text-Authentic} is much smaller than the others, further demonstrating that the spoken language texts generated by \textsc{PGen} are closer to the domain of \textsc{Text-Authentic}. These demonstrate the effectiveness and generalizability of the proposed \textsc{PGen} approach.


\begin{table}[t!] 
    \centering
    \setlength{\tabcolsep}{3pt}
    \begin{tabular}{l | ccc}
        \toprule
       \bf Data & \bf JS$\downarrow$ \\
        \midrule 
        \textsc{Text-PGen}\quad\quad~~ vs. \textsc{Text-Authentic}  & 0.01 \\
        \textsc{Text-Selected} ~ vs. \textsc{Text-Authentic} & 0.18 \\
        \textsc{Text-General}\quad vs. \textsc{Text-Authentic} & 0.26\\
        \bottomrule
    \end{tabular}
    \caption{The JS divergence between different types of spoken language texts for Phoenix2014T dataset.}
    \label{tab:divergence}
\end{table}

\begin{table}[t!] 
    \centering
    \begin{tabular}{l | r r}
        \toprule
        \bf Test Data & \bf In-domain & \bf General \\
        \midrule
        \textsc{Text-Authentic}  & 99.38\%  & 0.62\% \\
        \textsc{Text-PGen}       & 98.60\%  & 1.40\%  \\
        \textsc{Text-Selected}  & 56.23\%  & 43.77\%   \\
        \textsc{Text-General}    & 0.31\%   & 99.69\% \\
        \bottomrule
    \end{tabular}
    \caption{The domain classification results of different spoken language texts on Phoenix2014T dataset. }
    \label{tab:sent_class}
\end{table}

\begin{table*}[t!] 
    \centering
    \setlength{\tabcolsep}{3pt}
    \scalebox{0.96}{
    \begin{tabular}{l | cccc | cccc}
        \toprule
         &  \multicolumn{4}{c|}{\bf Dev Set} &  \multicolumn{4}{c}{\bf Test Set} \\
         & BLEU-1 & BLEU-2 & BLEU-3 & BLEU-4 & BLEU-1 & BLEU-2 & BLEU-3 & BLEU-4 \\
        \midrule
        \multicolumn{9}{c}{\bf \em Phoenix2014T}\\
        \citet{camgoz2018neural} & 44.40 & 31.83 & 24.61 & 20.16 & 44.13 & 31.47 & 23.89 & 19.26 \\
        \citet{camgoz2020sign}   & \bf 50.69 & \bf 38.16 & 30.53 & 25.35 & \bf 48.90 & 36.88 & 29.45 & 24.54 \\
        \citet{yin2020better}    & 49.05 & 36.20 & 28.53 & 23.52 & 47.69 & 35.52 & 28.17 & 23.32  \\
        Transformer 
         & 43.05 & 32.57 & 25.50 & 20.81 & 43.71 & 33.40 & 26.45 & 21.73\\
        ~~~~+ Scaling BT(General)   & 43.52 & 32.42 & 25.30 & 20.62 & 42.68 & 32.06 & 25.09 & 20.40 \\
        ~~~~+ Scaling BT(Selected) & 44.20 & 33.02 & 25.86 & 21.06 & 44.29 & 33.25 & 25.97 & 21.09 \\
        
         ~~~~+ Scaling BT(PGen) & 
        48.68 & 37.94 & \bf 30.58 & \bf 25.56 & 48.30 & \bf 37.59 & \bf 30.32 & \bf 25.54  \\

        \midrule
        \multicolumn{9}{c}{\bf \em ASLG-PC12}\\
        \newcite{yin2020better} & 92.67 & 88.72 & 85.22 & 81.93 & 92.88 & 89.22 & 85.95 & 82.87 \\
        Transformer 
         & 91.85 & 87.53 & 83.73 & 80.19 & 92.04 & 88.07 & 84.56 & 81.25 \\
        ~~~~+ Scaling BT(General)  & 91.59 & 87.43 & 83.81 & 80.45 & 91.97 & 88.19 & 84.89 & 81.79 \\
        
        ~~~~+ Scaling BT(PGen)  & \bf 93.23 & \bf 88.91 & \bf 85.63 & \bf 82.04 & \bf 93.51 & \bf 89.74 & \bf 86.55 & \bf 83.35 \\
        \midrule
        \multicolumn{9}{c}{\bf \em CSL-Daily}\\
        Transformer
        & 49.63 & 35.62 & 25.52 & 18.64 & 49.41 & 35.57 & 25.55 & 18.72 \\
        ~~~~+ Scaling BT(General)   & 54.66 & 39.80 & 29.23 & 21.78 & 54.07 & 39.34 & 28.99 & 21.75 \\

        ~~~~+ Scaling BT(PGen)  & \bf 60.48 & \bf 46.92 & \bf 36.95 & \bf 29.72 & \bf 60.21 & \bf 46.76 & \bf 36.90 & \bf 29.75 \\

        \bottomrule
    \end{tabular}
    }
    \caption{Gloss-to-text translation performance on Phoenix2014T, ASLG-PC12 and CSL-daily. "+ Scaling BT(PGen)" represents that training data is increased by 40 times with BT, in which the monolingual is generated by our PGen approach. }
    \label{tab:main}
\end{table*}

\paragraph{Domain Classifier.}
The word frequency distribution only characterizes one aspect of the domain. More features, for example, the styles of texts, can not be explicitly modeled. Thus, we follow~\citet{du2020adversarial} to train a domain classifier to distinguish the in-domain and the general domain data with the consideration of all potential features implicitly. We perform a binary classification task with equal examples (i.e., train/valid as 7086/519) from \textsc{Text-General} and \textsc{Text-Authentic}, respectively. 
To train the domain classifier, we finetune the German BERT\footnote{\url{https://huggingface.co/bert-base-german-cased}} on the above dataset~\cite{sun2019fine}. Specifically, we use the German BERT to encode the input sentence and feed the \texttt{[CLS]} token vector as a reasonable sentence embedding to the domain discriminator. For testing, we also sample the same number of examples from the other three spoken language text corpora as the test set of \textsc{Text-Authentic}~(i.e., 642) and predict their domains (i.e., general or in-domai).

The results are listed in Table~\ref{tab:sent_class}. We observe that the domain classifier successfully predicts the true labels of \textsc{Text-Authentic} and \textsc{Text-General}, indicating the significant domain differences between sign language spoken texts and general spoken language texts. As for \textsc{Text-PGen}, the examples are categorized into authentic texts with very high accuracy~(i.e., 98.60\%), while the value is much lower for \textsc{Text-Selected}~(i.e., 56.23\%). These results again demonstrate that our \textsc{PGen} approach can produce better in-domain spoken language texts than the compared methods.

\subsection{Sign Language Gloss Translation}
\label{sec:sign_lagnauge_glsss_translation}

We perform extrinsic evaluation for the proposed \textsc{PGen} by applying back-translation with the generated spoken language texts to the sign language gloss translation task. For the different SLT tasks in \ref{tab:data}, we adopt the corresponding German\footnote{\url{https://huggingface.co/dbmdz/german-gpt2}},  Chinese\footnote{\url{https://huggingface.co/uer/gpt2-chinese-cluecorpussmall}} and English\footnote{ \url{https://huggingface.co/gpt2} } GPT-2 models to generate in-domain spoken language texts with \textsc{PGen}. 

For the gloss-to-text translation task, we follow \citet{yin2020better} to train a Transformer model with 2 encoder layers and 2 decoder layers. 
For back-translation, we first finetune the mT5 pre-trained model on the authentic text-gloss parallel data and then use it to translate the collected spoken language texts (sampled from \textsc{Text-PGen}, \textsc{Text-Selected} or \textsc{Text-General}) into glosses. Following~\citet{wu2019exploiting}, we train the Transformer model on the combination of the authentic and synthetic parallel data, and then finetune it on the authentic gloss-to-text parallel data. 

For evaluation, we follow previous studies~\cite{camgoz2018neural,camgoz2020sign, zhou2021improving} to evaluate the performance of gloss-to-text translation with BLEU score~\cite{papineni2002bleu}, ROUGE-L~\cite{lin2004rouge}, and METEOR~\cite{banerjee2005meteor} scores. Specifically, we report the BLEU-1,2,3,4 scores to reflect the translation quality at different phrase levels. 

Table~\ref{tab:main} lists the main results of the gloss-to-text translation performance on the Phoenix2014T, ASLG-PC12 and CSL-daily datasets. By scaling the BT synthetic data to 40 times of the authentic parallel data with the different types of spoken text data, \textsc{Text-PGen} (i.e., ``+~Scaling BT(PGen)'') improves the performance over the baseline Transformer model significantly and consistently (e.g., up to 11.03 BLEU-4 points on CSL-daily), while \textsc{Text-General} and \textsc{Text-Selected}
only improve slightly or even hurt the performances~(e.g., down to 1.33 BLEU-4 points on Phoenix2014).
We also report the translation performance in terms of ROUGE-L and METEOR in  Appendix~\ref{sec:performance}. These demonstrate the effectiveness and generalizability of the proposed \textsc{PGen} approach.

\begin{table*}[t!] 
    \centering
    \setlength{\tabcolsep}{3pt}
    \begin{tabular}{l | cccc  | cccc}
        \toprule
         &  \multicolumn{4}{c|}{\bf Text-to-Gloss Dev Set} &  \multicolumn{4}{c}{\bf Gloss-to-Text Dev Set} \\
         & BLEU-1 & BLEU-2 & BLEU-3 & BLEU-4 & BLEU-1 & BLEU-2 & BLEU-3 & BLEU-4 \\
        \midrule
        Transformer & -- & -- & -- & -- &  43.05 & 32.57 & 25.50 & 20.81 \\
        ~~~~+ BT Model SC       & 57.72 & 39.41 & 27.86 & 19.66 & 44.80 & 34.33 & 27.47 & 22.86 \\
        ~~~~+ BT Model mT5    &  \bf 59.78 & \bf 43.70 & \bf 32.58 & \bf 25.01 & \bf 46.87 & \bf 35.95 & \bf 28.59 & \bf 23.54\\
        \bottomrule
    \end{tabular}
    \caption{Translation performance with different back-translation models on text-to-gloss and gloss-to-text tasks.}
    \label{tab:BT_performance}
\end{table*}

\section{Analysis}
\label{sec:analysis}
To gain a deeper understanding on our \textsc{PGen} approach, we conduct extensive analyses in terms of ablation study and translation outputs.

\subsection{Ablation Study}
\label{sec:abaltion}

We conduct three ablation studies regarding the prompt length, the back-translation model and the scale of synthetic data on the Phoenix2014T dataset. We introduce more details as below:

\begin{figure}[t]
    \centering
	\begin{minipage}{0.80\linewidth}
		\vspace{3pt}

		\centerline{\includegraphics[width=\textwidth]{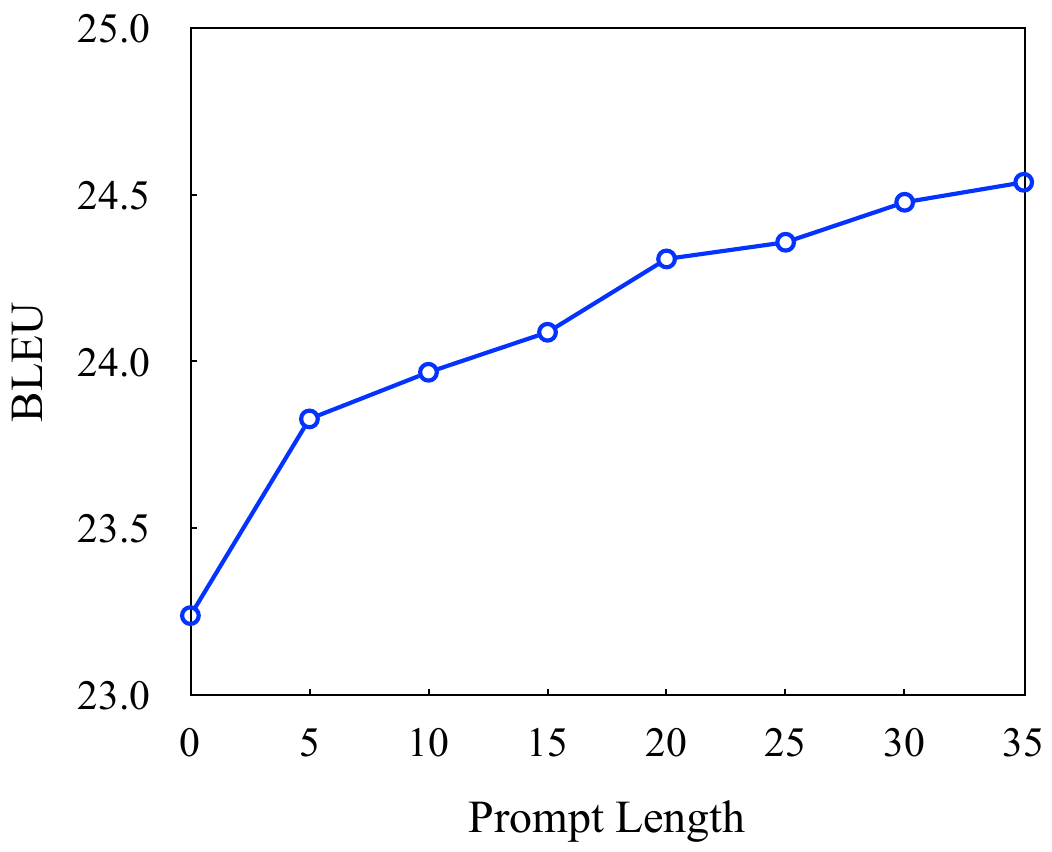}}
	\end{minipage}
	
	\caption{Gloss-to-text translation performance with respect to different prompt sizes. The X-axis represents the number of sentences in one prefix, while the Y-axis represents the BLEU-4 score.}
	\label{fig:Prefix_length}
\end{figure}

\paragraph{Prompt Length.}
We first investigate the impact of the prompt length $k$ on the gloss-to-text translation task, which decides how many sentences are concatenated as prompts for generation via \textsc{PGen}. Specifically, we increase $k$ from 0 to 35 to induce the PLM to generate in-domain spoken language texts with a data size 5 times of the authentic parallel data. 
Then, we perform back-translation with the generated spoken language texts on the gloss-to-text translation task, individually. 
The results are shown in Figure~\ref{fig:Prefix_length}, in which we observe that the performance of gloss-to-text translation constantly improves with the increase of prompt length.
This is because a larger prompt length can provide more domain signals so as to encourage the generation of higher-quality spoken language texts with closer domain and higher diversity (see Appendix~\ref{sec:diversity}).
However, the larger prompt length requires more computation memory, slowing down the generation process. Therefore, we set the prompt length to 20 throughout the work for a good tradeoff between the generation quality and the computation costs.

\paragraph{Back-Translation Model.}

In~\cref{sec:approach-BT}, we state that the quality of BT models heavily affects the performance of final models. To validate this claim, we compare two back-translation models:
a Transformer model trained from scratch~(i.e., ``+~BT Model SC'') and the finetuned mT5~\cite{xue2020mt5} model~(i.e., ``+~BT Model mT5'').
As shown in the left of Table~\ref{tab:BT_performance}, the finetuned mT5 model produces higher-quality pseudo parallel data according to BLEU scores on the validation set of the text-to-gloss translation~(i.e., back-translation) task. Consequently, the performance of gloss-to-text translation is considerably improved when synthesizing data by ``+~BT Model mT5'', which reconfirms our claim. Therefore, throughout this work, we adopt the finetuned mT5 model for back-translation.

\paragraph{Scale of Synthetic Data.}

In~\cref{sec:approach-PGen}, we show the potential of \textsc{PGen} in generating large-scale in-domain spoken language texts. Let us recap Figure~\ref{fig:valid-bleu-vs-scale}, where we increase the scale of spoken language texts used for back-translation. We observe that scaling the spoken language texts generated by \textsc{PGen} can improve the performance of the gloss-to-text translation task consistently while that by retrieval degrades the performance. It suggests that our approach can scale the BT technique to play its maximum effect for gloss-to-text translation, which has not been achieved in previous studies.

\subsection{Translation Output}

We conduct further analyses to understand how the proposed approach improves the gloss-to-text translation quality. Specifically, we analyze the translation outputs of Phoenix2014 in Table~\ref{tab:main} by the \texttt{compare-mt}\footnote{\url{https://github.com/neulab/compare-mt}} toolkit in terms of word frequency and sentence length.

\paragraph{Words Frequency.}
Previous study~\cite{fadaee-monz-2018-back-2} shows that the back-translation improves the translation performance by improving the low-frequency word predictions. Meanwhile, our analyses in~\cref{sec:domain_text_generation} suggest that the spoken langugae texts generated by our \textsc{PGen} shows a similar word frequency distribution as \textsc{Text-Authentic}. We wonder how such consistency benefits the prediction of low-frequency words in gloss-to-text translation.
Specifically, we first categorize the vocabulary into three groups based on the word frequency in the training data, including High: frequency $\in [2000, +\infty)$; Medium: frequency $\in [100, 2000)$; Low: frequency $\in (0, 100]$. Then, we utilize \texttt{compare-mt} to calculate the prediction accuracy of target words in the test set with respect to the three groups. 

Table~\ref{tab:freq} lists the results for different models with scaling back-translation (e.g. 40 times synthetic data ). As seen, scaling back-translation with in-domain spoken language texts of \textsc{Text-PGen} improves the prediction of words in all three groups, especially for low-frequency words.
However, the situation is much different for \textsc{Text-Selected} and \textsc{Text-General} such that they bring little improvement for high-frequency words and inversely harm the performance on low-frequency words.
It indicates that back-translation becomes ineffective when the domains of spoken language texts and authentic parallel data are mismatched, which implies the importance of our \textsc{PGen} approach.

\begin{table}[t!] 
    \centering
    \setlength{\tabcolsep}{3pt}
    \begin{tabular}{l ccc}
        \toprule
        \multirow{2}{*}{\bf Data} & \multicolumn{3}{c}{\bf Word Frequency} \\
        \cmidrule(lr){2-4}
        &  Low & Medium & High  \\ 
        \midrule
        \textsc{Text-Authentic}   & 28.86 & 49.94 & 58.27  \\
        ~~~~+ \textsc{Text-PGen}   & 33.23\cellcolor{jgreen!43} & 52.98\cellcolor{jgreen!30} & 60.17\cellcolor{jgreen!19} \\
        ~~~~+ \textsc{Text-Selected}   & 24.65\cellcolor{jred!42} & 49.87\cellcolor{jred!1} & 58.58\cellcolor{jred!3}  \\
        ~~~~+ \textsc{Text-General}   & 25.45\cellcolor{jred!34} & 49.12\cellcolor{jred!8} & 58.12\cellcolor{jred!2} \\
        \bottomrule
    \end{tabular}
    
    \caption{Prediction accuracy~(F1 score) of target words in the test set with respect to word frequency. As the higher F1 score, the better, we mark the improvement by \colorbox{jgreen!50}{green} and degradation by \colorbox{jred!50}{red} background. 
    }
    \label{tab:freq}
\end{table}

\paragraph{Sentence Length.}
We investigate the translation quality of examples with varied lengths, which can be biased during generating or retrieving spoken language texts.
Similar to word frequency, we also categorize the examples of test set into three groups based on the sentence length, including Long: $(20, +\infty)$ tokens; Medium: $(10, 20]$ tokens; Short: $(0, 10]$ tokens.

\begin{table}[t!] 
    \centering
    \fontsize{10}{11}\selectfont
    \begin{tabular}{l ccc}
        \toprule
        \multirow{2}{*}{\bf Data} & \multicolumn{3}{c}{\bf Sentence Length} \\
        \cmidrule(lr){2-4}
        & Short & Medium & Long \\ 
        \midrule
        \textsc{Text-Authentic}   & 21.73 & 22.61 & 10.00  \\
        ~~~~+ \textsc{Text-PGen}   & 21.87\cellcolor{jgreen!1} & 27.29\cellcolor{jgreen!46} & 20.15\cellcolor{jgreen!101} \\
        ~~~~+ \textsc{Text-Selected}   & 16.24\cellcolor{jred!55} & 23.93\cellcolor{jgreen!13} & 12.80\cellcolor{jgreen!28}  \\
        ~~~~+ \textsc{Text-General}   & 16.68\cellcolor{jred!50} & 23.97\cellcolor{jgreen!13} & 9.47\cellcolor{jred!5} \\
        \bottomrule
    \end{tabular}
    
    \caption{Translation quality~(BLEU score) of examples in the test set with respect to sentence length. 
    }
    \label{tab:length}
\end{table}

Table~\ref{tab:length} lists the corresponding results. Clearly, long sentences are more difficult to translate~\cite{zheng2020improved} and our \textsc{PGen} can particularly improve the translation quality of medium and long sentences. In contrast, the other methods show little improvement on medium and long sentences and degrade the performance on short sentences significantly.
This demonstrates the better stability of our approach regarding the distribution of sentence length over the compared methods.

\section{Related Work}

\paragraph{Sign Language Gloss Translation.}
SLGT translates sign gloss to spoken language texts, which has attracted more attention in recent years with the development of neural machine translation~(NMT). 
For example, \citet{camgoz2018neural} released the PHOENIX14T and for the first time proposed a neural SLT model to translate from spatial representations or sign glosses. 
Recent studies attempt to improve both SLR and gloss-to-text translation for the better performance of SLT.
\citet{yin2020better} proposed the STMC-Transformer network~\cite{zhou2020spatial} to improve SLR and exploited Transformer for gloss-to-text translation.
\citet{camgoz2020sign} formulated SLR and gloss-to-text translation in the multi-task form while \citet{li2020tspnet} explored the hierarchical structure for learning sign video representations.
More recently, multi-cue characteristics of sign language have also been utilized for improving SLT~\cite{camgoz2020multi, zhou2021spatial, kan2022sign}.
Different from these work, we improve SLT by focusing on the gloss-to-text translation task in the perspective of spoken language generation.

\paragraph{Data Augmentation.}
Data augmentation has been proposed and proven valuable and effective in machine translation research~\cite{sennrich2016improving,zhang2016exploiting,wang2018switchout,jiao2020data,jiao2021self}. To address the data scarcity issue in gloss-to-text translation, there have been studies on producing synthetic gloss-text pairs for data augmentation.
One stream is to extract discrete phrases from natural texts based on linguistic rules~\cite{moryossef2021data}.
Another is to adopt BT technique~\cite{sennrich2016improving} to generate glosses from natural texts by a pretrained text-to-gloss translation model. However, the limited in-domain spoken language texts prevent BT from playing its maximum effect for the gloss-to-text translation task. While we may collect in-domain texts from public websites as \citet{zhou2021improving}, it is unreliable due to both the requirement for domain knowledge and the accessibility of websites.
Our approach exploits the knowledge and generalization capability of large PLMs to produce in-domain texts with prompt learning.
While large PLMs have been successfully applied for text generation in NLP~\cite{kumar2020data,chintagunta2021medically}, we craft the prompts to theoretically guarantee that we can produce large-scale in-domain texts based on the small-scale original spoken language texts data.

\section{Conclusion}

In this paper, we propose the \textsc{PGen} approach to produce large-scale in-domain monolingual texts based on the original small-scale texts of the gloss-text parallel data. With \textsc{PGen}, we scale back-translation and achieve significant and consistent improvements on three benchmark datasets for SLGT task. Extensive analyses suggest that our approach generates monolingual texts with similar linguistic properties as the original monolingual texts, thus outperforms the compared methods in terms of both low-frequency word prediction and long sentence translation. 
Future work includes exploring ChatGPT for sign language translation task by using proper prompts~\cite{jiao2023chatgpt}.

\bibliography{anthology,custom}
\bibliographystyle{acl_natbib}

\appendix

\label{sec:appendix}

\newpage
\clearpage

\section{Appendix}
\subsection{Model Training}
\label{sec:hyper}

\begin{table*}[t] 
    \centering
    \begin{tabular}{l c c c c}
        \toprule
        \multirow{2}{*}{\bf Parameter} & \multicolumn{2}{c}{\bf Text-to-Gloss} &  \multicolumn{2}{c}{\bf Gloss-to-Text} \\
          & \bf Transformer  &  \bf mT5   &  \bf Pretrain & \bf Finetune\\
        \midrule
         encoder-layers  & 2  & 12                   & 6 & 6 \\
         decoder-layers  & 2  & 12                   & 6 &  6\\
         learning rate   & $7 \cdot 10^{-4}$ & $7 \cdot 10^{-4}$     & $7 \cdot 10^{-4}$ & $7 \cdot 10^{-4}$ \\
         learning rate scheduler & inverse\_sqrt & inverse\_sqrt & inverse\_sqrt & inverse\_sqrt\\
         Adam $\beta$    &  (0.9, 0.98) & - &(0.9, 0.98) &  (0.9, 0.98)\\
         warmup-updates  & 2000 & - & 4000 & 1000\\
         label-smoothing & 0.1  & 0.1       & 0.1 & 0.1\\
         dropout         & 0.3     & 0.1    & 0.1 & 0.5 \\
         batch-size      & 2048    & 2048    & 2048 & 2048 \\
        \bottomrule
    \end{tabular}
    \caption{Hyperparameters of translation models.}
    \label{tab:hype_para}
\end{table*}

We perform extrinsic evaluation for the proposed \textsc{PGen} by applying back-translation with the generated spoken language texts to the sign language gloss translation task. For the Chinese and English SLT tasks, we adopt the corresponding English\footnote{ \url{https://huggingface.co/gpt2} } and Chinese\footnote{\url{https://huggingface.co/uer/gpt2-chinese-cluecorpussmall}} GPT-2 models to generate in-domain spoken language texts with \textsc{PGen}. 

For the gloss-to-text translation task, we follow \citet{yin2020better} to train a Transformer model with 2 encoder layers and 2 decoder layers. 
We use Adam~\cite{kingma2015adam} with $\beta = (0.9, 0.98)$ and $\epsilon = 10^{-6}$ for optimization. We adopt the warm-up learning rate scheduler, which linearly increases from $1.0\times10^{-4}$ to a peak of $5.0\times 10^{-4}$ within 2000 steps, and then decays with the inverse square root schedule. The dropout rate is 0.3 and the label smoothing is 0.1. 

For back-translation, we first finetune the mT5 pre-trained model on the authentic text-gloss parallel data and then use it to translate the collected spoken language texts (sampled from \textsc{Text-PGen} or \textsc{Text-Selected}) into glosses. Following~\citet{wu2019exploiting}, we train the Transformer model on the combination of the authentic and synthetic parallel data, and then finetune it on the authentic gloss-to-text parallel data. 

Table~\ref{tab:hype_para} presents the hyper-parameters of different transformer models used in this work.

\begin{figure}[h]
    \centering
	\begin{minipage}{0.80\columnwidth}
		\vspace{3pt}
        
		\centerline{\includegraphics[width=\textwidth]{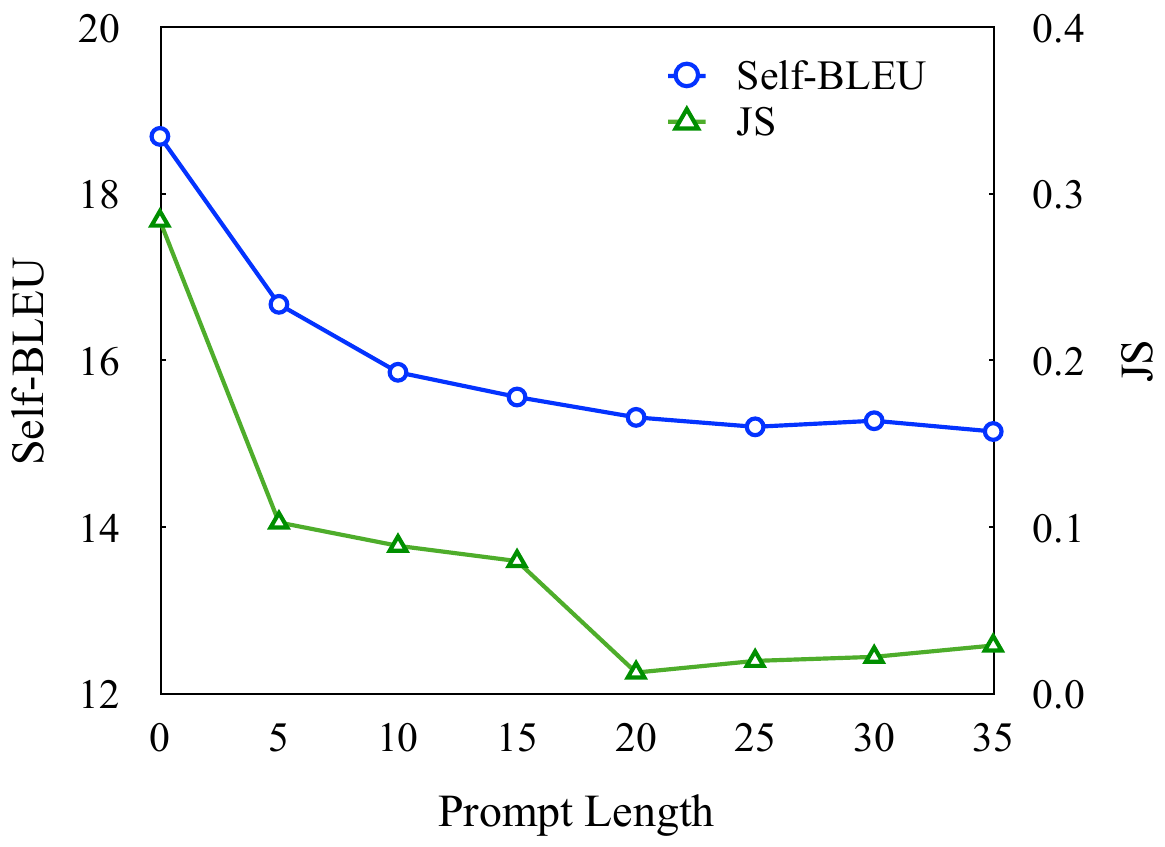}}
          
	\end{minipage}
	\caption{Diversity and domain closeness of sign language texts generated by \textsc{PGen} measured by Self-BLEU and JS with respect to different prompt sizes.}
	\label{fig:diversity_vs_prompt_length}
\end{figure}

\subsection{Diversity and Quality}
\label{sec:diversity}
Figure~\ref{fig:diversity_vs_prompt_length} shows the diversity and domain closeness of sign language texts generated by \textsc{PGen} measured by Self-BLEU and JS scores with respect to prompt length. Lower Self-BLEU~\cite{zhu2018texygen} and JS scores indicate higher diversity and closer domain, respectively.

\subsection{Intrinsic Analsys Results for ASLG-
PC12 and CSL-Daily Datasets}
\label{sec:ex-intrinsic}
We extend the intrinsic analyses to both ASLG-PC12 and CSL-Daily datasets. 
Tabel~\ref{tab:ex-divergence} shows the JS divergence between different types of spoken language texts on the two SLT datasets. As seen, the JS divergence from \textsc{Text-PGen} to \textsc{Text-Authentic} is much smaller than \textsc{Text-General} to \textsc{Text-Authentic}. Table~\ref{tab:ex_sent_class} lists the domain classification results on the ASLG-PC12 and CSL-Daily datasets. The results indicate the significant domain differences between sign language spoken texts and general spoken language texts, and our \textsc{PGen} approach can produce in-domain spoken language texts. 

Clearly, the results on both ASLG-PC12 and CSL-Daily datasets are consistent with that in section~\ref{sec:domain_text_generation}, which demonstrates the effectiveness and generalizability of the proposed
\textsc{PGen} approach.

\begin{table*}[ht] 
    \centering
    \begin{tabular}{l | ccc | ccc}
        \toprule
       \bf Data & \bf ASLG-PC12 & \bf CSL-Daily \\
        \midrule 
        \textsc{Text-PGen}\quad\quad~~ vs. \textsc{Text-Authentic}  & 0.02 & 0.08\\
        \textsc{Text-General}\quad vs. \textsc{Text-Authentic} & 0.18 & 0.14\\

        \bottomrule
    \end{tabular}
    \caption{The JS divergence results on the ASLG-PC12 and CSL-Daily datasets.}
    \label{tab:ex-divergence}
\end{table*}

\begin{table*}[ht] 
\begin{minipage}[c]{0.5\textwidth}
    \begin{tabular}{l | r r}
        \toprule
        \bf Test Data & \bf In-domain & \bf General \\
        \midrule
        \textsc{Text-Authentic}  & 98.63\%  & 1.37\% \\
        \textsc{Text-PGen}       & 99.44\%  & 0.56\%  \\
        \textsc{Text-General}    & 1.58\%   & 98.42\% \\
        \bottomrule
    \end{tabular}
\caption*{ASLG-PC12}
\end{minipage}
\begin{minipage}[c]{0.5\textwidth}
    \begin{tabular}{l | r r}
        \toprule
        \bf Test Data & \bf In-domain & \bf General \\
        \midrule
        \textsc{Text-Authentic}  & 99.86\%  & 0.14\% \\
        \textsc{Text-PGen}       & 98.57\%  & 1.43\%  \\
        \textsc{Text-General}    & 0.24\%   & 99.76\% \\
        \bottomrule
    \end{tabular}
\caption*{CSL-Daily}
\end{minipage}

\caption{The domain classification results on the ASLG-PC12 and CSL-Daily datasets. }
\label{tab:ex_sent_class}
\end{table*}

\subsection{Other Metrics}
\label{sec:performance}

\begin{table*}[ht] 
    \centering
    \setlength{\tabcolsep}{3pt}
    \begin{tabular}{l  cc  cc}
        \toprule
         &  \multicolumn{2}{c}{\bf  Dev Set } &  \multicolumn{2}{c}{\bf Test Set} \\
         & ROUGE-L & METEOR & ROUGE-L & METEOR\\
        \midrule
        \multicolumn{5}{c}{\bf \em Phoenix2014T}\\
        \citet{camgoz2018neural}  & 46.02 & - & 45.45 & -\\
        \citet{yin2020better}     & 47.36 & \bf 46.09 & 46.58 & 44.85 \\
        Transformer  & 47.77 & 43.46 & 47.48 & 42.36  \\
         ~~~~+ Scaling BT(General) & 46.43 & 42.35 & 46.12 & 42.24 \\
         ~~~~+ Scaling BT(Selected) & 48.16 & 43.35 & 47.95 & 42.96 \\
         ~~~~+ Scaling BT(PGen) & \bf 50.89 & 45.50 & \bf 49.25 & \bf 44.78 \\
         
        \midrule
        \multicolumn{5}{c}{\bf \em ASLG-PC12}\\
        \citet{yin2020better} & 82.41 & 95.93 & 95.87 & 96.46 \\
        Transformer  & 91.45 & 92.85 & 94.74 & 95.30 \\
         ~~~~+ Scaling BT(General) & 91.65 & 92.95 & 94.96 & 95.50 \\
        ~~~~+ Scaling BT(PGen)  & \bf 94.82 & \bf 96.79 & \bf 96.43 & \bf 96.79 \\
         
        \midrule
        \multicolumn{5}{c}{\bf \em CSL-Daily} \\
        Transformer   & 40.94 & 23.78 & 40.87 & 23.53 \\
         ~~~~+ Scaling BT(General) & 55.76 & 36.28 & 44.64 & 35.62 \\
        ~~~~+ Scaling BT(PGen) & \bf 62.31 & \bf 50.48 & \bf 60.54 & \bf 50.35\\

        \bottomrule
    \end{tabular}
    \caption{Gloss-to-text translation performance on Phoenix2014T, ASLG-PC12 and CSL-daily. "+ Scaling BT(PGen)" represents that training data is increased by 40 times with BT, in which the monolingual is generated by our PGen approach.  }
    \label{tab:ROUGE}
\end{table*}

Table~\ref{tab:ROUGE} presents the ROUGE-L and METEOR scores of the gloss-to-text translation performance on the Phoenix2014T, ASLG-PC12 and CSL-daily datasets.
\end{document}